\documentclass[sigconf]{acmart}

\makeatletter
\def\@ACM@checkaffil{
    \if@ACM@instpresent\else
    \ClassWarningNoLine{\@classname}{No institution present for an affiliation}%
    \fi
    \if@ACM@citypresent\else
    \ClassWarningNoLine{\@classname}{No city present for an affiliation}%
    \fi
    \if@ACM@countrypresent\else
        \ClassWarningNoLine{\@classname}{No country present for an affiliation}%
    \fi
}
\makeatother
\AtBeginDocument{%
  }

\setcopyright{acmlicensed}
\copyrightyear{2018}
\acmYear{2018}
\acmDOI{XXXXXXX.XXXXXXX}
\acmConference[Conference acronym 'XX]{Make sure to enter the correct
  conference title from your rights confirmation email}{June 03--05,
  2018}{Woodstock, NY}
\acmISBN{978-1-4503-XXXX-X/2018/0}

\usepackage{booktabs}
\usepackage{multirow}
\begin{document}

\title{OCP: Orthogonal Constrained Projection for Sparse Scaling in Industrial Commodity Recommendation
}




\author{Chen Sun}

\email{sunchen59@jd.com}
\affiliation{%
  }

\author{Beilin Xu}
\email{xubeilin.1@jd.com}
\affiliation{%
  }

\author{Boheng Tan}
\email{tanboheng.1@jd.com}
\affiliation{%
  }

\author{Jiacheng Wang}
\email{wangjiacheng.52@jd.com}
\affiliation{%
  }

\author{Yuefeng Sun}
\email{sunyuefeng6@jd.com}
\affiliation{%
  }

\author{Rite Bo}
\email{borite.1@jd.com}
\affiliation{%
  }

\author{Ying He}
\affiliation{%
  \institution{JD.com}
  \city{Beijing}
  \country{China}}
\email{heying.146@jd.com}

\author{Yaqiang Zang}
\authornotemark[1]
\affiliation{%
  \institution{JD.com}
  \city{Beijing}
  \country{China}}
\email{zangyaqiang.1@jd.com}

\author{Pinghua Gong}
\affiliation{%
  \institution{JD.com}
  \city{Beijing}
  \country{China}}
\email{gongpinghua1@jd.com}
\authornote{Corresponding author.}



\begin{abstract}
In industrial commodity recommendation systems, the representation quality of Item-Id vocabularies directly impacts the scalability and generalization ability of recommendation models. A key challenge is that traditional Item-Id vocabularies, when subjected to sparse scaling, suffer from low-frequency information interference, which restricts their expressive power for massive item sets and leads to representation collapse. 
To address this issue, we propose an Orthogonal Constrained Projection method to optimize embedding representation. By enforcing orthogonality, the projection constrains the backpropagation manifold, aligning the singular value spectrum of the learned embeddings with the orthogonal basis. This alignment ensures high singular entropy, thereby preserving isotropic generalized features while suppressing spurious correlations and overfitting to rare items. Empirical results demonstrate that OCP accelerates loss convergence and enhances the model's scalability; notably, it enables consistent performance gains when scaling up dense layers. Large-scale industrial deployment on JD.com further confirms its efficacy, yielding a 12.97\% increase in UCXR and an 8.9\% uplift in GMV, highlighting its robust utility for scaling up both sparse vocabularies and dense architectures.
\end{abstract}

\ccsdesc{Information systems~Recommender systems}

\keywords{Recommender systems, orthogonal constraint, sparse scaling}


\maketitle

\section{Introduction}
Modern industrial recommendation systems process massive and heterogeneous feature sets to model complex user behaviors. Among all input features, item representation quality is a core factor behind final performance. In practice, item features usually include categorical IDs, Item-IDs, and semantic IDs (SIDs).

Item-IDs and categorical entities are typically mapped to high-dimensional vectors through trainable lookup tables. During training, each input ID retrieves its embedding, and the embedding parameters are optimized end-to-end together with the recommendation objective. This paradigm is widely adopted in models such as YouTube DNN\cite{dnn}, DeepFM\cite{deepfm}, and DIN\cite{din}.  
SIDs are ordered codewords obtained by encoding item text/content into dense vectors with a pre-trained encoder and then quantizing them\cite{tiger, rqvae, vqvae, hou2025generating}. They allow semantically similar items to share statistical strength, which is especially helpful for cold-start and long-tail scenarios.

\begin{figure*}[t]
    \centering
    \includegraphics[width=1\linewidth]{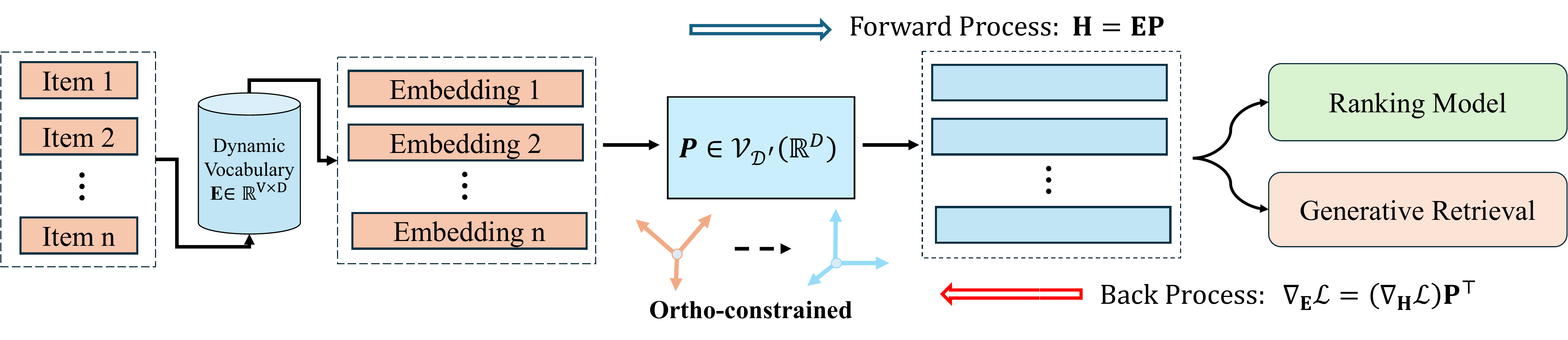}
    \caption{ 
        Overview of OCP.
    }
    \label{fig:main}
\end{figure*}

To capture subtle preference signals (for example, user affinity for a specific brand-model pair), models still need fine-grained Item-ID features. This shift from coarse to fine-grained representation has pushed industrial systems into the \textit{sparse scaling} regime\cite{store, msn, xiong2026llatte}, where vocabulary size $V$ expands from millions to billions to cover long-tail inventory.  
However, sparse scaling introduces a major optimization issue. User-item interactions are highly skewed, so frequent items receive dense updates while most long-tail Item-IDs are rarely updated. Over time, optimization concentrates on a few dominant directions, causing effective rank degradation in embedding space, i.e., embedding collapse.

For the purpose of improving representation quality, enforced orthogonality, such as orthogonal codebook regularization and orthogonal residual quantization\cite{guo2024embedding, liu2025muon, lee2025cross, yin2026dual, xu2025mmq_v2, xu2025mmq}, is adopted in recommendation system. Orthogonal Rotation Transformation is employed to rotate the subspace spanned by low-cardinality static features including categorical IDs\cite{store}. 
Meanwhile, Item-ID remains the finest-grained identifier and has the highest cardinality in production systems. Without proper constraints, models may overfit by memorizing sparse long-tail IDs rather than learning transferable patterns.  
To address this, we propose OCP (\textbf{O}rthogonal \textbf{C}onstrained \textbf{P}rojection) for Item-ID embedding optimization. Our contributions are:

\begin{itemize}
\item We characterize embedding collapse in sparse scaling and use Singular Entropy (SE) to quantify representation isotropy. The analysis explains how long-tail sparsity induces dimensional redundancy.

\item We constrain the projection matrix on the Stiefel manifold\cite{bernstein2025manifolds}, which preserves gradient geometry and stabilizes the singular-value distribution of embeddings.

\item We validate OCP with both offline experiments and large-scale online deployment. The online model achieves +12.97\% UCXR and +8.95\% GMV, showing effectiveness under both sparse (vocabulary) and dense (architecture) scaling.

\end{itemize}




\section{Methodology}

In this section, we first formalize sparse scaling (Section \ref{pre}), then analyze embedding collapse in sparse vocabularies (Section \ref{emb_collap}), and finally present OCP with its theoretical properties (Section \ref{ortho}).

\subsection{Preliminary
}
\label{pre}
In industrial recommendation systems, representation quality sets the performance ceiling. Category IDs provide stable global semantics with low cardinality; SIDs derived from clustering or pre-trained encoders capture intermediate-granularity semantics; Item-IDs provide the finest-grained identity with very high cardinality.

Formally, let $\mathbf{E} \in \mathbb{R}^{V \times D}$ denote the Item-ID embedding matrix and let $\mathbf{P} \in \mathbb{R}^{D \times D'}$ be a projection matrix for downstream interaction layers:

\begin{equation}
    \mathbf{H} = \mathbf{E}\mathbf{P}.
\end{equation}

While Sparse Scaling endows the model with the capacity to represent long-tail items, it simultaneously introduces a fundamental challenge rooted in distributional imbalance. The majority of entries in $\mathbf{E}$ correspond to extremely low-frequency Item-IDs, which receive insufficient gradient updates due to sparse interaction data. More critically, under the conventional unconstrained parameterization of $\mathbf{P}$, the absence of directional constraints on gradient flow triggers a pathological phenomenon known as dimensional collapse: the effective rank of the learned representation space degrades as optimization concentrates update signals along a small subset of dominant dimensions. 


\subsection{Embedding Collapse in Scaled Models}
\label{emb_collap}

Empirical evidence suggests that sparse scaling often leads to embedding collapse\cite{guo2024embedding}. Despite the high-dimensional embedding space $\mathbb{R}^D$, the learned embedding vectors $\mathbf{E}$ tend to reside in a much lower-dimensional manifold. This dimensionality redundancy implies that the model fails to utilize its increased capacity, resulting in stagnant or even degraded performance as the model scales.




To quantify collapse severity and information density, we use Singular Entropy (SE)\cite{liu2025muon}. For $\mathbf{E} \in \mathbb{R}^{V \times D}$ with singular values $\{\sigma_1, \sigma_2, \dots, \sigma_k\}$ ($k=\min(V,D)$), define energy $e_i=\sigma_i^2$ and normalized energy $p_i=e_i/\sum_{j=1}^{k}e_j$. SE is defined as:
\begin{equation}
\text{SE}(\mathbf{E}) = -\frac{1}{\log(k)} \sum_{i=1}^{k} p_i \log(p_i).
\end{equation}
A higher SE (closer to 1) indicates a more isotropic and better-utilized embedding space; a lower SE indicates concentration in a few dominant dimensions and thus more severe collapse.

\subsection{Orthogonal Constrained Projection Layer}
\label{ortho}

\subsubsection{Orthogonal Constraint on Stiefel Manifold}

In industrial recommendation architectures, the high-cardinality Item-ID embedding matrix $\mathbf{E} \in \mathbb{R}^{V \times D}$ is mapped to an interaction space via a projection layer $\mathbf{P} \in \mathbb{R}^{D \times D'}$:
$\mathbf{H} = \mathbf{E} \mathbf{P},$
where $V$ is vocabulary size, $\mathbf{H} \in \mathbb{R}^{V \times D'}$ is the projected representation, and $D$ is embedding dimension. Under sparse scaling, $\mathbf{E}$ often occupies a low-dimensional subspace, i.e., $\text{rank}(\mathbf{E}) \ll \min(V, D)$.

To avoid unconstrained optimization that may further distort gradient geometry, we optimize $\mathbf{P}$ on the Stiefel manifold:
\[
\mathcal{V}_{D'}(\mathbb{R}^D) = \{ P \in \mathbb{R}^{D \times D'} \mid P^\top P = I_{D'} \}.
\]

 
We apply a projection-based update:
\[
\mathbf{P} \leftarrow \text{QR}(\mathbf{P} - \eta \nabla_{\mathbf{P}}\mathcal{L}).
\]
This mechanism ensures forward-backward consistency: In the forward pass, $\mathbf{P}$ acts as an isometry, embedding high-cardinality items into a latent space without losing relative geometric relationships. In the backward pass, as derived in (\ref{eq:loss}), the gradient $\nabla_{\mathbf{E}}\mathcal{L}$ inherits the singular value spectrum of the upstream task gradient, effectively forcing the model to explore a broader representation space for infrequent items by ensuring gradient's orthogonality.


\begin{equation}
\label{eq:loss}
    \nabla_{\mathbf{E}}\mathcal{L} = \frac{\partial \mathcal{L}}{\partial \mathbf{H}} \frac{\partial \mathbf{H}}{\partial \mathbf{E}} = (\nabla_{\mathbf{H}}\mathcal{L}) \mathbf{P}^\top,
\end{equation}
where $\nabla_{\mathbf{H}}\mathcal{L} \in \mathbb{R}^{V \times D'}$ represents the upstream gradient from the interaction layers.

\subsubsection{Analysis of Singular Value Preservation}

Since $\mathbf{P}$ is column-orthogonal ($\mathbf{P}^\top\mathbf{P}=\mathbf{I}$), it acts as an isometry on the projected subspace. Therefore, for $i \le \text{rank}(\nabla_{\mathbf{H}}\mathcal{L})$, the non-zero singular values are preserved during backpropagation:
\begin{equation}
    \sigma_i(\nabla_{\mathbf{E}}\mathcal{L}) = \sigma_i((\nabla_{\mathbf{H}}\mathcal{L}) \mathbf{P}^\top) = \sigma_i(\nabla_{\mathbf{H}}\mathcal{L}).
\end{equation}

This alignment ensures that the gradient flow does not suffer from further rank erosion caused by the projection weights. By preventing the "squeezing" of the gradient manifold, OCP forces the embeddings of low-frequency items to be updated along diverse orthogonal directions, effectively "stretching" the representation space and mitigating embedding collapse.

\section{Experiments}

\subsection{Experimental Setup}

We evaluate OCP on both a generative retrieval model and a ranking model in JD.com's production environment.

\subsubsection{Baseline}
For generative retrieval, we use OxygenREC\cite{hao2025oxygenrec} as baseline, which retrieves items using three-level SIDs.  
For ranking, we use an online production model from a large e-commerce scenario at JD.com as baseline for both offline and online evaluations.
\subsubsection{Evaluation Metric}

For OxygenREC, we report Top1 hit@k ($k=1,2,3$), which measures whether the model correctly predicts the first-level SID, the first two SID levels, and the full three-level SID, respectively.  
For ranking, we report AUC and GAUC.

\subsubsection{Training and Evaluation Protocol}
For fair comparison, baseline and OCP models share the same backbone architecture, data split, optimizer family, and training budget (same number of steps).  
Unless otherwise specified, vocabulary settings follow each experiment configuration. For offline experiments, OCP is the only additional component over baseline.  
Offline metrics are computed on held-out data, and online results are reported from controlled A/B testing against the production baseline.

\subsubsection{Reproducibility Notes}
To improve reproducibility, we report key setup details used throughout the paper: (1) ranking experiments are trained on four months of interaction logs; (2) the SE analysis uses a $d=64$ Item-ID embedding slice; (3) sparse scaling experiments vary vocabulary from 1e to 10e ($1\text{e}=10^8$). The generative model was trained for 196K steps on a dataset of 6.4 billion samples, while the ranking model underwent 720K steps of training on 2.9 billion samples. 

\subsection{Overall Performance}
\begin{table}[]
\caption{Performance of OCP on OxygenREC with 0.7B and 3.2B parameters. In this paper, 1e denotes $10^8$. OCP consistently improves Top1 hit@k ($k=1,2,3$).}
\begin{tabular}{@{}cccccc@{}}
\toprule
Params                & Method         & Vocab size & Hit1  & Hit2 & Hit3 \\ \midrule
\multirow{2}{*}{0.7B model} & base      & 5e    & 23.42\% & 7.40\%  & 4.57\%  \\
                      & OCP        & 5e    & \textbf{23.94\%} & \textbf{7.58\%} & \textbf{4.66\%} \\ \midrule
\multirow{2}{*}{3.2B model} &  base     & 5e    & 23.50\% & 7.60\% & 4.73\% \\
                      & OCP        & 5e   & \textbf{23.93\%}  & \textbf{7.84\%} & \textbf{4.90\%} \\ \bottomrule
\end{tabular}
\label{clk_top1}
\end{table}

Table \ref{clk_top1} presents the overall prediction performance in our generative retrieval model. The sparse Item-ID vocabulary size is set to 0.5 billion. Compared to baseline, proposed method obtained higher Top1 hit@k (k=1,2,3) rate on both 0.7B and 3.2B model. After scaling up the model density, QR orthogonalization continued to yield improvements in evaluation metrics. 
Unlike baselines lacking vocabulary optimization, our proposed method employs orthogonal space constraints to refine the embedding manifold. This approach effectively enhances the model's capacity for feature generalization while suppressing the overfitting of memorized noise. Consequently, our method achieves a significant gain in predictive performance by maintaining a high-rank representation space.

\begin{figure}
    \centering
    \includegraphics[width=1\linewidth]{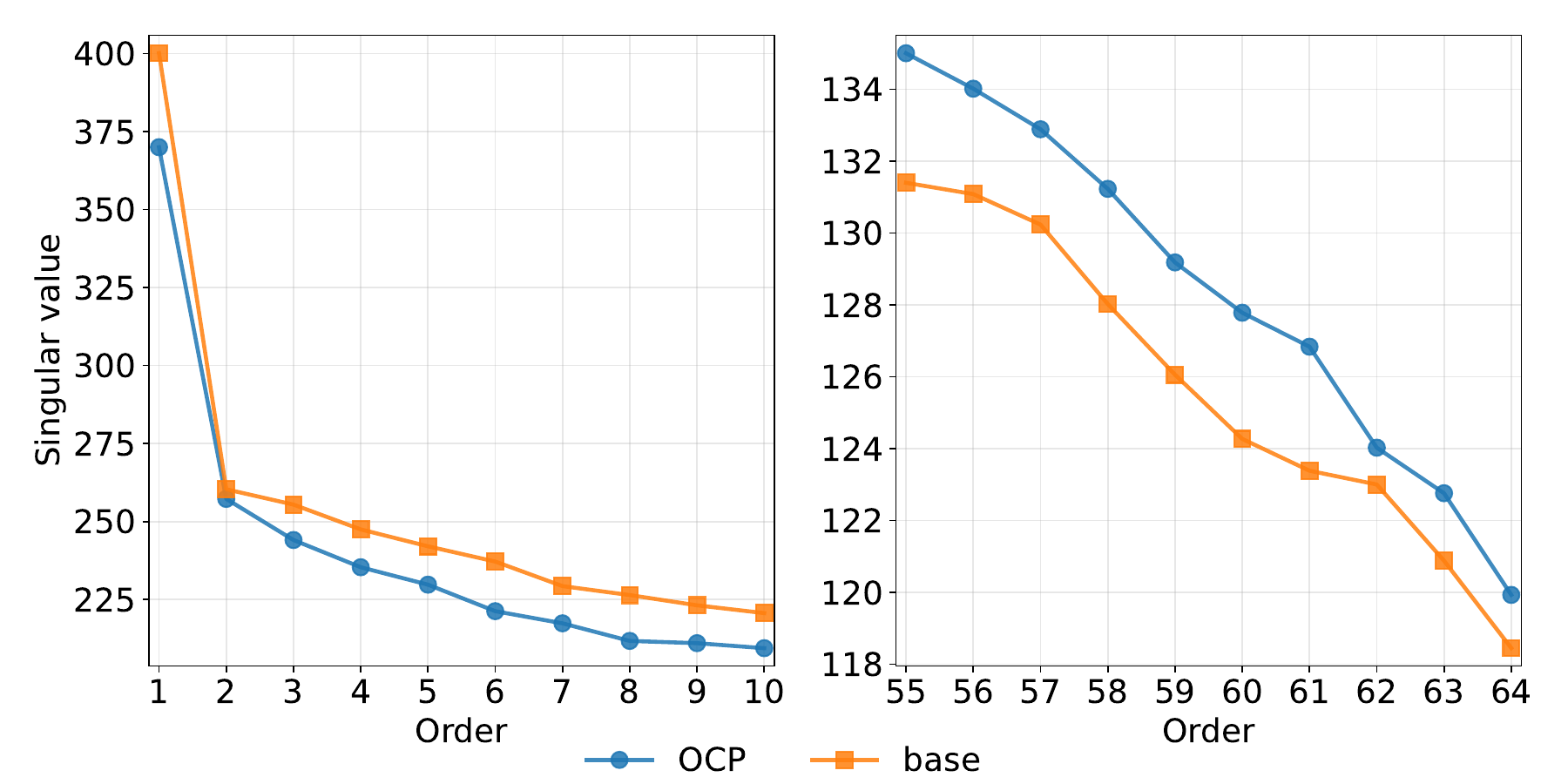}
    \caption{Comparison of singular value distributions. Orders 1--10 and 55--64 represent the top-10 and bottom-10 singular values of the $d=64$ embedding matrix, respectively. 
    }
    \label{fig:sv}
\end{figure}

\subsection{Singular Entropy Study}
We investigate effect of OCP on geometric stability of Item-ID embeddings using a transformer-based ranking model trained on four months of interaction data from JD.com. The vocabulary size is 1 billion.

To evaluate representation quality, we analyze singular values of the embedding matrix. As shown in Figure \ref{fig:sv}, OCP reduces overly dominant top singular values while strengthening bottom singular values, indicating a more \textit{isotropic} space and less collapse.

We then conduct stratified analysis on (i) all items, (ii) bottom 80\% low-frequency items, and (iii) top 20\% high-frequency items.  
As shown in Table \ref{tab:se}, OCP improves SE in all three groups, indicating two effects: it preserves non-zero singular values under column-orthogonal projection and regularizes over-dominant frequent-item directions.

\begin{table}[]
\caption{Singular Entropy (SE) results. Higher entropy indicates model's enhanced capacity for sparse representation.}
\begin{tabular}{llll}
\hline
         & All data        & Bottom 80\%      & Top 20\%        \\ \hline
base     & 0.9725          & 0.9646          & 0.9705          \\
OCP      & 0.9794 (+0.71\%) & 0.9715 (+0.72\%) & 0.9787 (+0.85\%) \\ \hline
\end{tabular}
\label{tab:se}
\end{table}

\subsection{Scaling Law Study with OCP}

In this section, we study scaling behavior on both OxygenREC and ranking models to evaluate OCP scalability.

\subsubsection{Dense Scaling} 
\begin{figure}
    \centering
    \includegraphics[width=\linewidth]{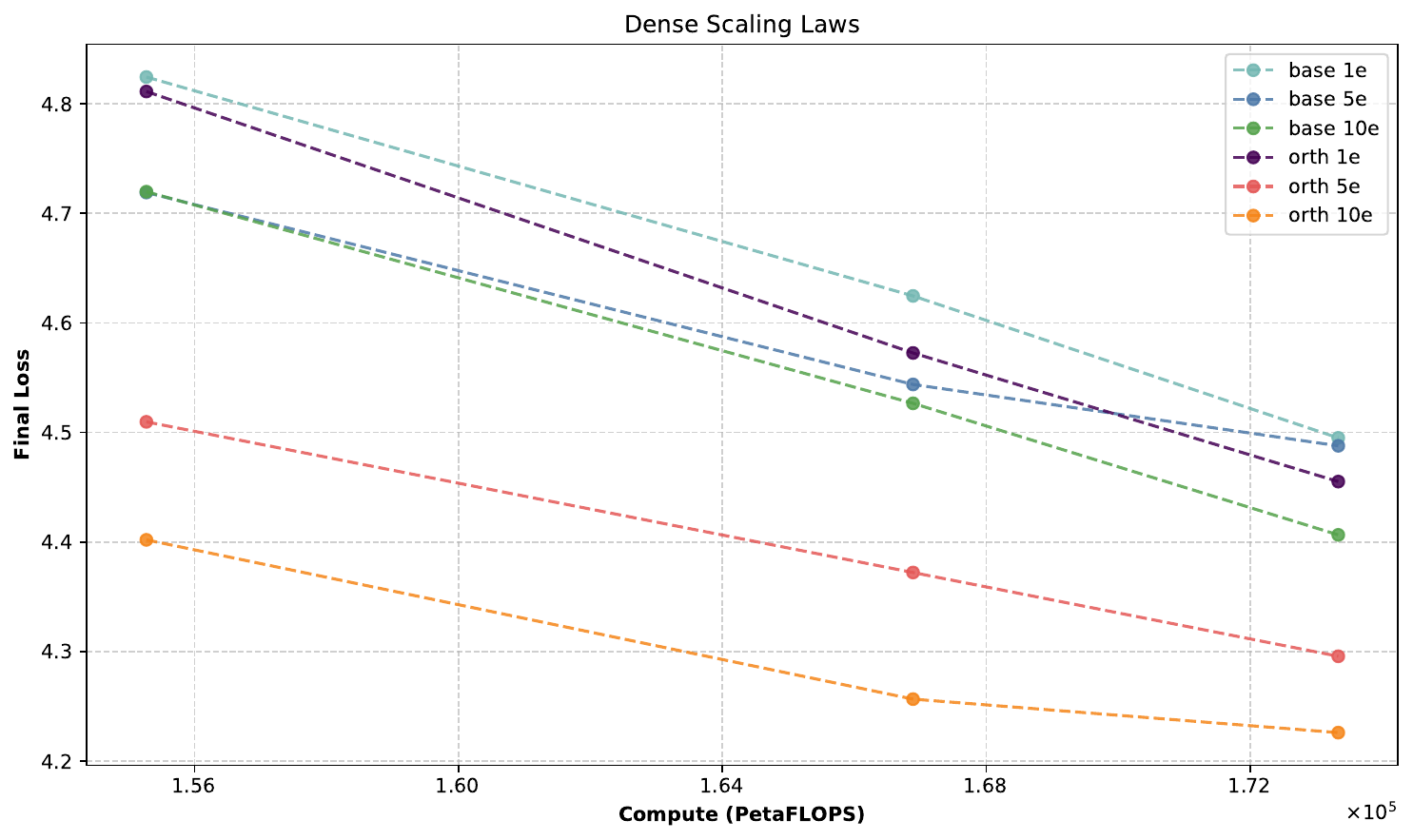}
\caption{Dense scaling law on OxygenREC with different model sizes. As parameters increase from 0.7B to 1.7B and 3.2B, OCP achieves lower final loss.}
    \label{fig:scaling_dense}
\end{figure}

As shown in Figure \ref{fig:scaling_dense}, increasing model parameters leads to lower loss and faster convergence under the same training steps.
The trend is consistent with standard scaling behavior in recommender training: more dense capacity improves fit quality when optimization remains stable.

\subsubsection{Sparse Scaling} 
\begin{figure}
    \centering
    \includegraphics[width=\linewidth]{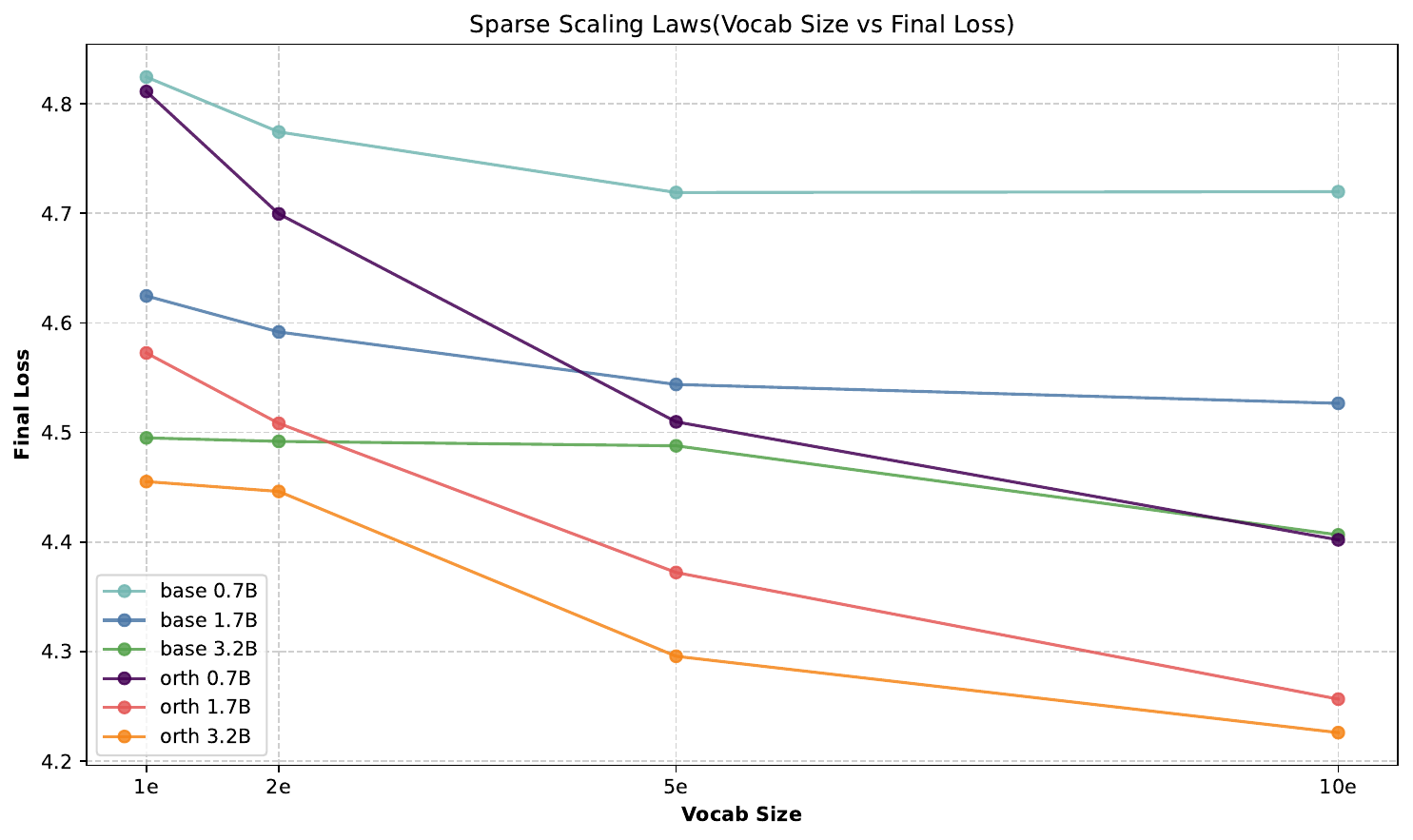}
\caption{Sparse scaling law on OxygenREC with different vocabulary sizes. As vocabulary increases from 1e ($10^8$) to 2e, 5e, and 10e, OCP
achieves a lower final loss.}
    \label{fig:scaling_sparse}
\end{figure}

As shown for OxygenREC in Figure \ref{fig:scaling_sparse}, under the same dense model size, expanding the Item-ID vocabulary
results in lower loss. We further run vocabulary ablations on the ranking model by tightening the access threshold (Table \ref{tab:thres}), which reduces vocabulary size and consistently degrades AUC/GAUC.  
These results indicate that preserving high-rank Item-ID representations is critical when scaling sparse vocabularies.

\subsubsection{What This Validates}
Taken together, the dense and sparse scaling results support two complementary conclusions:  
(1) OCP does not block gains from increasing dense model capacity;  
(2) OCP makes large sparse vocabularies more useful by preserving representation quality rather than letting extra IDs collapse into noisy low-rank directions.

\subsubsection{Discussion on Overhead}
OCP introduces a QR-based retraction step for the projection matrix. In our setting, this cost is small compared with the overall training cost dominated by large embedding lookups and transformer interaction layers.  
Given its consistent gains across retrieval and ranking tasks, OCP provides a favorable trade-off between additional computation and quality improvement in production.

\begin{table}[]
\caption{Performance under different vocabulary access thresholds. A higher threshold means stricter entry and a smaller vocabulary. Metrics include AUC/GAUC for click (clk) and order (ord).}
\begin{tabular}{ccccc}
\hline
access threshold & clk AUC & clk GAUC & ord AUC & ord GAUC     \\ \hline
15         & 0.7701 & 0.6281 & 0.8158 & 0.6491 \\ 
10          & 0.7738 & 0.6376 & 0.8179 &0.6519 \\
5         & 0.7761 & 0.6404 & 0.8195 & 0.6539 \\
3         & \textbf{0.7764} & \textbf{0.6432} & \textbf{0.8199} & \textbf{0.6562} \\ \hline

\label{tab:thres}
\end{tabular}
\end{table}

\subsection{Online Experiments}
We conducted a 5-day online A/B test on JD.com. With OCP, the Item-ID vocabulary scales from 178 million to 1 billion. Combined with a multi-layer self-attention architecture, OCP captures finer long-tail representations and improves key business metrics over production baseline: +12.97\% UCXR, +13.07\% order volume, and +8.95\% GMV.

These online gains are directionally consistent with offline findings: better singular-spectrum health and stronger long-tail representation quality translate to measurable business impact.

\subsection{Limitations and Future Work}
Our current study focuses on a single orthogonality mechanism (QR retraction) and fixed Item-ID vocabulary construction pipelines.  
Future work includes: (1) comparing alternative manifold optimizers with lower overhead, (2) integrating dynamic vocabulary growth and pruning, and (3) analyzing OCP under stronger distribution shifts (e.g., seasonal and cold-start bursts) with longer online evaluation windows.

\section{Conclusion}

In this paper, Orthogonal Constrained Projection (OCP) is introduced to improve Item-ID embedding quality under sparse scaling. By constraining the projection matrix on the Stiefel manifold, OCP preserves healthier gradient geometry and mitigates embedding collapse.  
Experiments on generative retrieval, ranking, and online deployment show that OCP consistently improves convergence and business metrics, and remains robust under both sparse and dense scaling.

\bibliography{arxiv}

\end{document}